\documentclass[10pt, a4paper]{article}

\usepackage[final]{lrec2026} 

\usepackage{times}
\usepackage{latexsym}

\usepackage[T1]{fontenc}

\usepackage[utf8]{inputenc}

\usepackage{microtype}

\usepackage{inconsolata}

\usepackage{color,soul}
\usepackage{booktabs}
\usepackage{multirow}
\usepackage{float}
\usepackage{enumitem}

\usepackage{graphicx}

\usepackage{tcolorbox}

\usepackage{listings}
\usepackage[table]{xcolor}
\usepackage{verbatim}
\usepackage{mdframed}
\usepackage{tabularx}

\title{Predicting Social Media User Actions: A Hybrid Approach for Common and Rare Behavior Prediction on Bluesky}

\name{Benjamin White, Anastasia Shimorina} 

\address{Orange Research, Lannion, France \\\\
         \{firstname.lastname\}@orange.com\\}

\abstract{
Understanding and predicting user behavior on social media platforms is crucial for content recommendation and platform design. While existing approaches focus primarily on common actions like retweeting and liking, the prediction of rare but significant behaviors remains largely unexplored. This paper presents a hybrid methodology for social media user behavior prediction that addresses both frequent and infrequent actions across a diverse action vocabulary. We evaluate our approach on a large-scale Bluesky dataset containing 6.4 million conversation threads spanning 12 distinct user actions across 25 persona clusters. Our methodology combines four complementary approaches: \textit{(i)} a lookup database system based on historical response patterns; \textit{(ii)} persona-specific LightGBM models with engineered temporal and semantic features for common actions; \textit{(iii)} a specialized hybrid neural architecture fusing textual and temporal representations for rare action classification; and \textit{(iv)} generation of text replies. Our persona-specific models achieve an average macro F1-score of 0.64 for common action prediction, while our rare action classifier achieves 0.56 macro F1-score across 10 rare actions. These results demonstrate that effective social media behavior prediction requires tailored modeling strategies recognizing fundamental differences between action types. Our approach achieved first place in the SocialSim: Social-Media Based Personas challenge organized at the Social Simulation with LLMs workshop at the Conference on Language Modeling (COLM 2025).
 \\ \newline  }

\begin{document}

\maketitleabstract
\section{Introduction}

Social media platforms have become central to modern communication, generating vast amounts of user interaction data that offer unprecedented insights into human behavior patterns. Understanding and predicting how users engage with content—whether they like, share, comment, or take other actions—has significant implications for both theoretical research in computational social science and practical applications in content recommendation, platform design, and user experience optimization.

The prediction of user behavior on social media platforms presents a complex challenge. Traditional approaches have primarily focused on the most frequent user actions, such as retweeting and liking, often employing hand-crafted features or network-based representations \cite{10.1109/SocialCom.2010.33, Yang2010UnderstandingRBA, 6137399}. However, the emergence of large language models (LLMs) has opened new possibilities for more sophisticated user behavior modeling through persona-based simulation and generation \cite{rossetti2024ysocialllmpoweredsocial, Trnberg2023SimulatingSM}.

Despite these advances, several key challenges remain unresolved. First, most existing work has concentrated on binary prediction tasks or a limited set of common actions, leaving the prediction of rare but potentially important behaviors (such as blocking, unfollowing, or content deletion) relatively unexplored \cite{Wu_Hu_Jia_Bu_He_Chua_2020}. Second, recent research suggests that smaller, specialized models may outperform large language models on action classification tasks \cite{qiu2025llmssimulatesocialmedia}. Third, the integration of temporal dynamics, user personas, and content semantics into a unified prediction framework remains an open challenge.

This work addresses these limitations by developing a comprehensive approach to social media user behavior prediction that handles both common and rare actions while incorporating rich temporal and semantic features. We present a hybrid methodology that combines lookup-based prediction for high-confidence cases, specialized tabular models for common actions, transformer-based architectures for rare action classification, and LLM-based generation of replies. 

Our key contributions are threefold: (1) We demonstrate that a portion of user behavior can be predicted through historical pattern matching; (2) We show that persona-specific tabular models using carefully engineered temporal and semantic features outperform transformer-based approaches for common action prediction; and (3) We develop a specialized hybrid neural architecture that effectively handles the class imbalance inherent in rare action prediction by fusing textual and temporal representations.

Our approach achieved first place in the SocialSim: Social-Media Based Personas challenge 2025\footnote{\url{https://sites.google.com/view/social-sims-with-llms/shared-task}, \url{https://www.kaggle.com/competitions/social-sim-challenge-social-media-based-personas/leaderboard}}, and it demonstrates how the strengths of several NLP approaches to social media text analysis can be combined to more accurately predict and simulate user behavior on a modern social media platform.

\section{Background}

\paragraph{Social Media User Behavior Prediction}

User behavior on social media platforms has been extensively studied, given its theoretical interest and industrial importance. With Twitter as a primary source of data, previous work has focused extensively on understanding and predicting retweeting behaviors \cite{10.1109/SocialCom.2010.33, Yang2010UnderstandingRBA} and using features based on the text message content \cite{6137399} or graph and network properties \cite{Sotiropoulos2019TwitterMancerPI}.

To improve behavior prediction, approaches which incorporate richer user-specific information have been developed. For example, \citet{FIRDAUS2021100165} combine users' tweeting and retweeting historical data to construct topic preferences and psychological profiles, and use these enriched persona features to improve retweet prediction F1 scores.

Clustering the potentially very large number of distinct users into a smaller number of persona groups has also been explored; in these approaches, users are grouped by behavior, demographic, or other similarity conditions, and then the user's membership in a particular persona cluster is incorporated in downstream prediction tasks. For example, \citet{Sun2023MeasuringTE} used persona information to predict user written replies to Twitter news headlines, with appropriate sentiment intensity and polarity.

To go beyond manual feature engineering, approaches using end-to-end deep learning have attracted much attention: an early approach \cite{zhang2016retweet} learnt joint representations of the tweet author, text, and the responding user to predict retweet behavior. Later approaches combine further actions and relationships obtained from the social media platform, such as network properties and message metadata, to improve prediction performance \cite{app122111216}.

While most research has focused on the more frequent actions of retweeting and liking, there has been comparatively less examination of other rarer specific actions. In this direction, \citet{Wu_Hu_Jia_Bu_He_Chua_2020} focused entirely on the unfollow action on the Weibo platform, finding the decision to unfollow another user to be more complex -- and thus requiring more sophisticated user and persona representations -- than e.g. the decision to follow another user.

\paragraph{Personas, LLMs, and Agentic Simulation}

With recent improvements in large language models (LLMs), research has explored the potential for using these systems for emulating social media platforms and their human users.

\citet{rossetti2024ysocialllmpoweredsocial} developed a LLM-powered digital twin of a social media platform, using agents characterized by age, interests, and personality traits, and based their modeling of agent activity on real data obtained from Bluesky.

\citet{Trnberg2023SimulatingSM} combined LLMs and agent-based modeling by creating realistic personas based on real demographic data, and demonstrate its applicability to studying complex user interaction scenarios.

Going further, \citet{touzel2024simulationsolvingsocietalscalemanipulation} built a sophisticated environment based on the Mastodon messaging app, enabling persona agents to take full control of their user accounts and emulate human interactions, and implemented a realistic longitudinal survey of the agents' political opinions and voting patterns.

\citet{Zhou2024KnowledgeBA} simulates user reactions on social media with an agentic approach, and incorporate persona information and grounding by implementing a dedicated persona module that informs the agent's planning and action decisions.

However, recent research has also identified several limitations of LLMs for simulating human behavior in general, outlining the beginning of a more rigorous science of persona generation \cite{li2025llmgeneratedpersonapromise}.

Using LLMs to simulate social media engagement including both subtasks of action prediction and written response generation,
has been explored recently by \citet{qiu2025llmssimulatesocialmedia}. Interestingly, the authors find that smaller fine-tuned BERT models outperform larger frontier LLMs on action classification tasks and furthermore that the LLMs exhibit particular sensitivity to the specific prompt context used for the classification task. However they find that LLMs perform well on the text generation subtask, and in particular that few-shot prompting with user information and historical examples improves the semantic alignment with reference tweets. Taken together, especially since their work explored a dataset with only 3 available actions (quote, rewrite, retweet), this recent work suggests that hybrid approaches to social media user simulation may still outperform purely LLM-based approaches, with the strength of LLM components lying in their ability to generate faithful text examples when augmented with relevant persona data obtained from other feature extraction approaches.

\section{Data and Task Description}
\label{sec:exploratorydataanalysis}
We describe our submission to the SocialSim challenge 2025, which distributed the persona-based social media dataset from Bluesky \cite{bückkaeffer2025textttblueprintsocialmediauser}. The dataset description paper was released only after the challenge had concluded. Therefore, below we present our exploratory data analysis as conducted for the challenge.

We use \textsc{BluePrint}, the persona-based social media dataset from Bluesky \cite{bückkaeffer2025textttblueprintsocialmediauser}. Bluesky is a modern microblogging social media platform with a similar interface and user experience to platforms such as X, allowing users to post messages and interact with messages and threads posted by other users. The public training/development dataset version available for the SocialSim 2025 challenge\footnote{\url{https://www.kaggle.com/competitions/social-sim-challenge-social-media-based-personas/data}} consists of 6,435,348/213,556 samples, and all results herein refer to evaluations performed on this development split (our results on the competition holdout private test set were similar). Each sample is a conversation thread taken from Bluesky, containing one or more messages in English written by the users who participated in the conversation. Each of the messages in a given thread contains the anonymized user ID, the content of the message, and a relative timestamp indicating when the message was posted to Bluesky.
In addition, the users in \textsc{BluePrint} are clustered into 25 different personas: user embeddings were obtained by pooling all individual posts, quotes, and replies authored by each user, and then these user embeddings were clustered to form 25 persona clusters. Inspection and TF-IDF analysis (visible in Table 10 of \citet{bückkaeffer2025textttblueprintsocialmediauser}) of the 25 obtained clusters reveals that they correspond to different behavioral and topic groups, with e.g. cluster \#22 corresponding mainly to creative workers who discuss art related topics. The final message in each thread of the \textsc{BluePrint} dataset contains the final user's persona cluster which allows for the analysis of persona specific social behavior in subsequent modeling and analysis.

The modeling task is to predict how a real user, knowing which of the 25 \textit{persona} clusters that user belonged to, would respond to the conversation i.e. firstly to predict the \texttt{action} that the user took, and secondly to generate the \texttt{text} that they wrote if applicable. Table~\ref{tab:examplefromtrainingdataset} shows an example from the dataset, and Table~\ref{tab:actiondistribution} provides the complete list of all 12 possible Bluesky user actions that appear in the training data.

\begin{table}[ht]
\centering
\begin{tabularx}{\linewidth}{l X}
\hline
Field & Value \\
\hline
first message time & 3,885,851 sec \\
first message text & Trump was filmed yesterday clearly ... \\
first user id & dd4724\\
second message time & 8,012,737 sec\\
second user id & 4ffd33 \\
second user cluster & 17 \\
\cellcolor{blue!25}second user action & \cellcolor{blue!25}FOLLOW \\
\cellcolor{blue!25}second message text & \cellcolor{blue!25}None \\
\hline
\end{tabularx}
\caption{
A dataset example where there are only 2 turns in the thread, so that the first user is the original poster and the second user from the persona cluster 17 is the one who is interacting with the post---in this case by deciding to follow the first user. Message times are measured in seconds, and are given relative to an unspecified start time---in this case the second user interacted 47 days after the first message was posted. We show in blue the fields that are to be predicted.}
\label{tab:examplefromtrainingdataset}
\end{table}

\paragraph{Persona Clusters} In total, 25 different persona clusters exist in the dataset.
From inspection of the text content of their messages, they mainly represent coherent behavioral groupings, such as scientific communities, sports fandoms, and political affiliations \cite{bückkaeffer2025textttblueprintsocialmediauser}.
Our manual reviewing of TF-IDF analyses on 25 clusters found that the primary distinction was whether clusters focused on political topics, with political clusters differing mainly in the intensity of political sentiment and country of interest, while non-political clusters were primarily centered on technology, art, or adult content.

\paragraph{Social Media Context} Public access to Bluesky opened in February 2024, leading to a rapid increase in users and the inclusion of very recent content, so content in the training data is potentially more up-to-date than that found in frontier LLMs. This influx likely consisted of entire communities with shared interests, resulting in an over-representation of certain viewpoints and a higher frequency of positive actions (e.g., LIKE; see Table~\ref{tab:actiondistribution}) compared to negative ones. Additionally, since most accounts are new, the high number of FOLLOW actions likely reflects users actively building their networks during this period.

\paragraph{Training Data Simplification} For the action prediction subtask, we retained only the conversations of length 2 because we noted that in the training dataset the actions for conversations of length 3 were always REPLY. Therefore we decided to systematically predict REPLY if the given sample contained more than 2 messages. This simplification reduced the number of REPLY actions from 47,798 (Table~\ref{tab:actiondistribution}) to 39,349.

\paragraph{Task Evaluation} The task performance is evaluated using F1-scores (both weighted and macro) for the prediction of the user's \texttt{action}, and a cosine similarity score for the \texttt{text} generation (i.e. how closely the model's predicted user \texttt{text} was to the text in the actual reply). Macro F1 was used as the primary metric in the SocialSim 2025 challenge, given the highly imbalanced dataset, to encourage performance across all actions including the rare actions. The cosine similarity is calculated using a text embedding model\footnote{\url{https://huggingface.co/intfloat/multilingual-e5-large}}.

\begin{table}[h!]
\centering
\begin{tabular}{|l|r|r|}
\hline
\textbf{Action} & \textbf{Count} & \textbf{Percentage} \\
\hline
follow        & 4,386,038 & 68.16\% \\
like          & 1,846,842 & 28.70\% \\
unfollow      &    97,804 &  1.52\% \\
\cellcolor{green!50}reply         &    47,798 &  0.74\% \\
quote         &    41,401 &  0.64\% \\
unlike        &    13,209 &  0.21\% \\
post\_update  &     1,014 &  0.02\% \\
repost        &       744 &  0.01\% \\
block         &       479 &  0.01\% \\
post\_delete  &        14 &  0.00\% \\
unblock       &         3 &  0.00\% \\
unrepost      &         2 &  0.00\% \\
\hline
\end{tabular}
\caption{Train dataset actions with counts and percentages. The REPLY action highlighted in green is generative---the action is accompanied by the user also inputting text.}
\label{tab:actiondistribution}
\end{table}

\paragraph{Task Simplification} We note here an important aspect of the dataset that simplifies the prediction task and may bias results or prevent generalization to all users of social media platforms: by the nature of the dataset construction, all the samples are associated with a definite user action. However, in the real setting of Bluesky, the choice of not taking any action is itself a possibility---indeed it may be the most frequent ``action'' across the entire platform as most users presumably do not interact with most messages. Therefore, the subset of messages that appear in the dataset are those which were for one reason or another able to elicit an interaction by a social media user, and may not be representative of the nature of generic social media messages.

\section{Methodology}
\label{sec:methodology}

\subsection{Prediction Pipeline Summary}

In this section, we outline our prediction pipeline and detail each step in subsequent sections. 

Each query in the dataset consists of a conversation with 1 or more messages with their text content, and raw relative timestamp data for each message. The query also contains the persona cluster of the user whose action we must predict in response to reading this conversation.

We first measure how many messages there are: if there are 3 or more messages, we predict the user action REPLY, based on our observation of this pattern in the training data as described in Section~\ref{sec:exploratorydataanalysis}.

For remaining samples with exactly 2 messages, we lookup the exact string of the first user's message in a lookup database of Bluesky messages. We determine whether we have sufficient information about all \textit{other users'} interactions with this message to perform a ``majority vote'' prediction: if many users from the same cluster have all responded in a very similar way, we take this most frequent action as our current prediction. Alternatively, if we do not have data from the same cluster, we also examine the entire user base but with a more selective threshold criterion (Section~\ref{subsec:lookupdatabase}).

Next, if the lookup database does not allow us to confidently make a prediction, we use a cluster-specific trained LightGBM model corresponding to the current user's persona cluster. This model will predict either FOLLOW, LIKE, or OTHER. If this model predicts FOLLOW or LIKE, we take this as the prediction. If the model predicts OTHER, it means that we believe that the user action was one of the ``rare actions'' (Section~\ref{subsec:tabularmodeldevelopment}).

If the previous step resulted in the LightGBM model predicting OTHER, we send the query to a specialized rare action classification transformer model, which will determine which of the 10 specific rare actions (BLOCK, UNFOLLOW, REPOST, etc.) to predict for this query (Section~\ref{subsec:rare-action-classif}).

Finally, after generating the action label predictions for the entire dataset we filter all those samples where the predicted action is REPLY. We send these conversations to an LLM in order to generate a prediction for the text that the user actually wrote (Section~\ref{subsec:textgener}).

\subsection{Repeated Messages Lookup Database}
\label{subsec:lookupdatabase}

We constructed a database of all first messages that appeared in our dataset and then aggregated, for each message, all of the observed action responses to that message on a cluster-by-cluster level.

\begin{table*}[t]
\centering
\begin{tabular}{|c|c|c|c|c|c|c|c|}
\hline
\textbf{Message} & \textbf{Cluster} & \textbf{Like} & \textbf{Follow} & \textbf{Block}  & \textbf{Unfollow}  \\
\hline
\multirow{3}{*}{``Here is a cute picture of a puppy''} & 0 & 9172 &  23 &  0 &  0    \\
                          & 1 & 254 & 0 & 0 &  0    \\
                          & 2 &  469 & 11 & 0 & 0    \\
\hline
\multirow{3}{*}{``I like pineapple on pizza''} & 0 & 0 & 0 & 710 & 2848     \\
                          & 1 & 946 & 671 & 20 & 178    \\
                          & 2 & 7894 & 541 & 0  & 0     \\
\hline
\end{tabular}
\caption{Artificial examples for the message lookup database step. Here we show 2 artificial examples and only 3 clusters and 4 actions, for clarity. We show different types of unanimity patterns as encountered in the real dataset: here the first message leads to essentially unanimous responses both within each cluster, and across the entire dataset. The second message (more divisive or controversial) leads to unanimous ``negative'' reactions in Cluster 0, unanimous ``positive'' reactions in Cluster 2, and heterogeneous reactions within Cluster 1.}
\label{tab:lookupdatabaseexamples}
\end{table*}

We show in Table~\ref{tab:lookupdatabaseexamples} some artificial (for pedagogical purposes) examples of entries in this lookup database. As with the illustrative examples in the table, real messages also vary in how unanimous the Bluesky users' responses are. Remarkably however, we found that many messages lead to quasi-unanimous response patterns within persona clusters or even in some cases at the entire global dataset level (i.e. when combining all responses from 25 persona clusters). This therefore suggested that it would be possible use a database lookup and majority-vote approach to generate predictions for such samples.

\begin{figure*}[tp]
  \centering
  \includegraphics[scale=0.4]{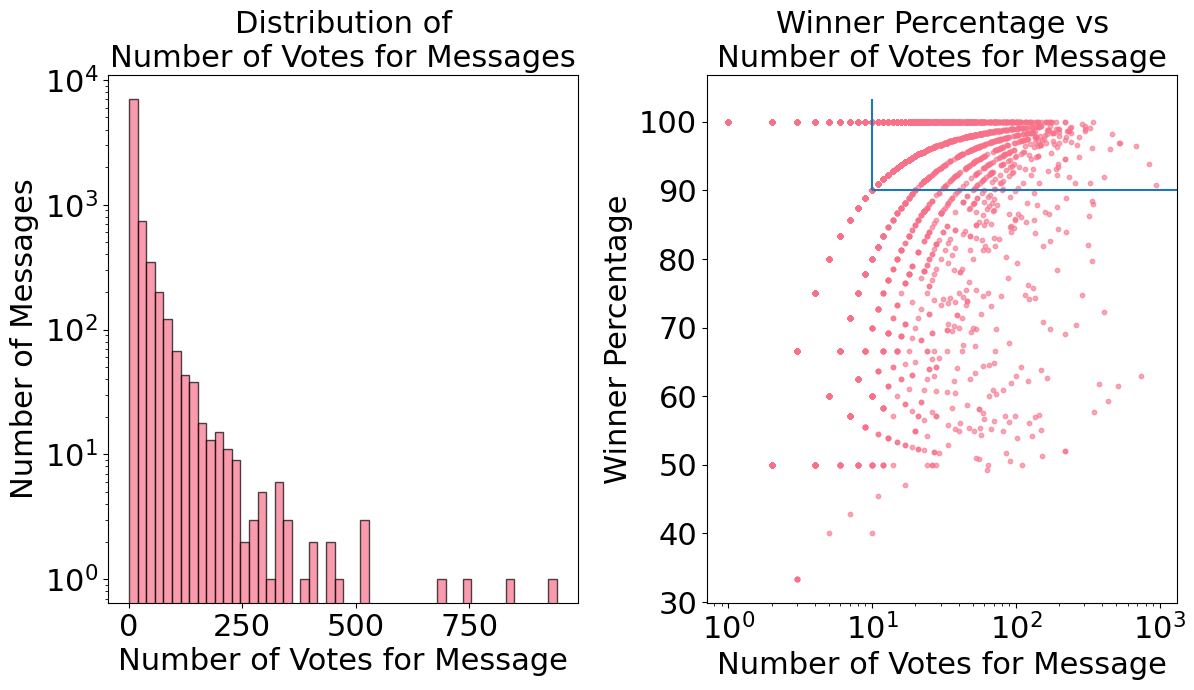}
  \caption{Sample of 10,000 messages from lookup database. In the right plot, the highlighted top-right rectangular region of interest contains messages with both a large number of votes (more than 10) and where the most frequent action has a high winner percentage (here more than 90\%) across all clusters.}
  \label{fig:lookupdatabase3plots}
\end{figure*}

We show in Figure~\ref{fig:lookupdatabase3plots} a sample of 10,000 messages from our lookup database. We refer to the number of times that different users have taken an action in response to a given message as the action ``votes'' for that message. For a given ``voting pattern'' (i.e. distribution of different actions) we refer to the most commonly taken action as the ``winner'' and its total vote frequency as the ``winner percentage''. Thus for example a message that 20 users have interacted with, leading to 18 LIKEs and 2 BLOCKs, would have the LIKE action as the winner, with a winner percentage of 90\%.

When examining the winner percentage for a given message, the total number of votes is an important variable: we consider that it is more significant if 200 users all respond identically to a message rather than if only 2 users respond identically.
In the right plot of the figure, the top-right region of the plot shows the messages with both a high total number of votes, and where the votes all tend to agree with one another leading to a high winner percentage; this is where we could be confident in making a prediction for a new user's action for a message based solely on other users' responses to that same message.

Since we recorded each message's action distribution as a function of each of the 25 clusters, we could decide to use solely the votes cast by other users from within the same shared persona cluster, or alternatively consider all the votes across all the clusters in the global database.
We defined 3 possible lookup strategies when using this message database for predictions: Cluster Specific (only take into account votes from other users in the same persona cluster), Global Fallback (if there aren't sufficiently many data points from within the same cluster, examine the global vote pattern across all users), and finally No Lookup Match (cases where even the Global Fallback strategy does not lead to identifying a ``winner action'' with a high winner percentage).

The specific thresholds of winner percentage and total votes to use for this lookup-based prediction are adjustable hyperparameters. We fixed a minimum number of total votes of 10 or more, and a ``winner percentage'' threshold of 85\% for the Cluster Specific strategy and of 90\% for the Global Fallback strategy. These thresholds were both lowered to 70\% for all messages where the ``winner action'' was one of the rare actions: this is because, after the subsequent main classifier model training steps (see later), we did not obtain any classifiers that achieved better than 70\% accuracy for the rare actions. Therefore we found it would always improve overall performance to just take majority-votes based on the lookup database rather than use the trained models.

\subsection{Tabular Model Development with Grouped Rare Actions}
\label{subsec:tabularmodeldevelopment}

We selected LightGBM \cite{ke2017lightgbm} as our modeling approach for the action classification task, after finding during initial exploration that deep learning approaches failed to attain good performances on this task: encoder language models (see Table~\ref{tab:lightgbmmodelsf1} for our baselines) attained macro-F1 scores of only 0.48 even after finetuning specialized social media models, while generative LLM approaches using in-context examples of the available actions on Bluesky produced even lower scores. 
\citet[Table~5]{bückkaeffer2025textttblueprintsocialmediauser} demonstrated similar results with LLM approaches, achieving F1 scores of 0.33 and showing only moderate gains from finetuning.

We developed a distinct LightGBM model for each of the 25 persona clusters: this approach allowed us subsequently to more clearly adjust model thresholds for classifier predictions, to study feature importance for each persona cluster.

In order to reduce the class imbalance, we replaced the 10 individual rare action labels (see Table~\ref{tab:actiondistribution}) with a synthetic grouped label containing all of these rare actions (label OTHER).
This therefore simplified the LightGBM training task to focus on only 3 possible classification labels: the 2 real and common actions (LIKE and FOLLOW) and 1 synthetic group (OTHER).

For LightGBM training\footnote{\url{https://github.com/microsoft/LightGBM}} we used class weights for the 3 action labels to handle imbalanced learning, weighting according to the inverse occurrence frequency of the 3 actions across the given cluster. We performed stratified 5-fold cross validation with hyperparameter tuning (number of estimators, learning rate, maximum depth, etc.), and for each of the 25 models we implemented threshold optimization using precision-recall curves.

\subsubsection{Feature Engineering}
We developed keyword, textual, and temporal features for LightGBM modelling.

\paragraph{Keyword Features}
Based on our exploratory data analysis,
most Bluesky messages were found to be well-written, thus enabling direct word matching rather than fuzzy string matching for any keywords.

We built an extensive keyword database, organized according to 11 primary topics (those identified during the persona cluster analysis, including politics, gaming, tech, etc.) and then specialized secondary subtopics.
We show in Table~\ref{tab:textkeywordcategories} the organization of this keyword database.

We applied all of these keyword features to our dataset: we recorded for each the total count of each keyword in order to have a measure of the ``intensity'' of each keyword - for example some messages contain the word ``Trump'' a dozen times, which we view as a different signal compared to a message where this word only appears once.

\begin{table}[t]
\centering
\begin{tabularx}{\linewidth}{X X p{0.4\linewidth}}
\toprule
Category & Subcategory & Keyword Features \\
\midrule
\multirow{3}{*}{Politics} & Candidates & biden, trump, harris \\
 & Issues & healthcare, taxes, border \\
 & Process & ballot, campaign, primary \\
\hline
\multirow{3}{*}{Bluesky} & Twitter & elon, leaving, bird app \\
 & Onboarding & new here, how do i \\
 & Community & invite codes, cozy  \\
\hline
\multirow{3}{*}{Gaming} & General & steam, xbox, switch \\
 & Streaming & twitch, vtuber \\
\bottomrule
\end{tabularx}
\caption{Selected examples of keyword features.}
\label{tab:textkeywordcategories}
\end{table}

\paragraph{Textual Features}
We constructed an additional set of various text features for basic text characteristics and metrics: total character count, word count, question mark count, hashtag count, etc. We also recorded the occurrence of various anonymization tags that occur in the dataset, such as \texttt{<USERNAME>} or \texttt{<URL>}. Some of these textual features can be seen in Table~\ref{tab:featureimportancelightgbm}.

\paragraph{Temporal Features}

For temporal features, we used the following information:
\begin{itemize}
    \item the time of the first message;
    \item the time of the second message;
    \item the time difference between the two messages;
    \item 7 categorical encodings derived from this time difference (e.g. \texttt{is\_immediate\_reply} if the time difference is less than 1 minute, etc.).
\end{itemize}

\subsection{Specialized Fine-grained Rare Action Classification}
\label{subsec:rare-action-classif}

We next built a dedicated model for handling samples that had been predicted as OTHER by LightGBM. This model's training objective was therefore to predict the specific individual rare action.

For this stage, due to the low frequency of the 10 rare actions across the entire dataset let alone within individual persona clusters, we trained a single rare action classification model using all available data rather than training 25 cluster-specific models.

Given that messages associated with rare actions would probably require more sophisticated semantic understanding and temporal features, we implemented a custom hybrid neural architecture to fuse textual representations with temporal behavior pattern representations.

We built a dual-branch neural network \cite{baltrušaitis2017multimodalmachinelearningsurvey}, containing both a text branch that takes an encoder language model to obtain a dense embedding for a given sample's text content as well as a temporal branch that uses dense layers to produce dense temporal embeddings. We then added an early fusion layer that combines both modalities, before sending this fused representation to a final classifier head.

For the text branch of our neural network, we used a model fine-tuned on Twitter\footnote{ \url{https://huggingface.co/cardiffnlp/twitter-roberta-base-emotion}, with the trained classifier head being ignored}, using the \texttt{CLS} token (a 768-dimensional embedding) as our text representation. This particular model has been trained on 154 million tweets from Twitter/X up until a December 2022 cut-off date \cite{antypas2023supertweeteval}, suggesting that the text content, modern writing style, and named entities might be particularly close to those from our Bluesky dataset.

For the temporal branch of our neural network, we augmented our existing timestamp features with several other features. Since the raw data contained relative timestamps (the actual message time and date was not specified), we created cyclical encodings corresponding to times within a day (even if the absolute value, i.e. \texttt{08:15am}, was not possible to determine, we could still assign a consistent relative timestamp from 0 to 86,400 seconds, i.e. 24h, to all samples in a cluster) and days within a week.

Our temporal branch module consists of a simple sequential module that scales all 12 temporal features as input, then passes through a 256-dimensional linear hidden layer trained with dropout and ReLU activation, followed by a second 128-dimensional linear hidden layer trained with dropout and ReLU activation to produce the temporal embeddings.

We experimented with different fusion architectures: early fusion by simply concatenating the 2 embedding types, late fusion by creating separate classifiers and using a learned weighting scheme, and finally a cross-attention fusion approach allowing temporal queries to attend to the text and using this attended text for classification. Across all hyperparameter searches we found that while attention fusion approaches could in some cases lead to higher performance, the early fusion approach was the most stable with consistently good results, so we retained it for our final model.

To handle class imbalance we implemented a focal loss \cite{lin2018focallossdenseobject} for our training criterion, using the class weights according to inverse occurrence frequency for the 10 rare actions across the training dataset, and a gamma value of 2.0 as its hyperparameter.

We found during development that a two-phase training strategy consisting of a warmup phase, where we froze the text encoder and trained the temporal/fusion layers for 2 epochs, followed by a fine-tuning phase, where we trained the architecture end-to-end for a further 3 epochs, improved overall model performance.

\subsection{Text Generation}
\label{subsec:textgener}

We used OpenAI GPT-4.1-mini\footnote{accessed: 28 August 2025} to generate reply messages by providing a conversation thread and requesting a suitable next response.

We implemented a characteristic Bluesky persona with a system prompt (``You are a politically liberal human social media user.''), since the site is used predominantly by liberal users; in the user prompt we described contextual information and dataset constraints (300 character limit, possibility of anonymized data in messages). See Appendix~\ref{subsec:appendix_prompts} for the whole prompt.

\section{Results}

Our prediction pipeline routed 0.96\% of samples using the simple rule-based approach on number of messages, 22.09\% of samples using the lookup database approach, 70.48\% of samples were classified as LIKE or FOLLOW directly by the sample's cluster-specific LightGBM model, with the remaining 6.47\% classified as OTHER by the LightGBM model and thus being sent to the custom trained rare action classifier. In total 6.60\% of all predictions were REPLY actions, that were therefore subsequently sent to the LLM stage to generate a plausible message text.

\begin{table}[h]
\centering
\setlength{\tabcolsep}{2.7pt}
\begin{tabular}{lccc}
    \textbf{} & \textbf{Min} & \textbf{Max} & \textbf{Avg} \\
    \midrule
    Majority Class Macro & 0.23 & 0.31 & 0.28\\
    Majority Class Weighted & 0.36 & 0.78 & 0.59 \\
    \hline
    RoBERTA-base Macro & - & - & 0.48 \\
    RoBERTA-base Weighted & - & - & 0.75 \\
    RoBERTA-Twitter Macro & - & - & 0.48 \\
    RoBERTA-Twitter Weighted & - & - & 0.75 \\
    \hline
    LightGBM Macro & \textbf{0.52} & \textbf{0.75} & \textbf{0.64} \\
    LightGBM Weighted & \textbf{0.81} & \textbf{0.88} & \textbf{0.83} \\
    \hline
    LightGBM Follow & 0.79 & 0.92 & 0.88 \\
    LightGBM Like & 0.57 & 0.86 & 0.73 \\
    LightGBM Other & 0.00 & 0.58 & 0.30 \\
\end{tabular}
\caption{
F1 scores for trained models on the simplified 3-label prediction task (FOLLOW, LIKE, OTHER). Baseline: majority class within each of the 25 clusters. Results include RoBERTa-base and RoBERTa-base-Twitter finetuned on 100,000 samples, and our LightGBM models per cluster, with min, max, and average scores across clusters.
}
\label{tab:lightgbmmodelsf1}
\end{table}

\begin{table*}[h!]
\centering
\begin{tabular}{|l r|l r|}
\hline
\textbf{Feature} & \textbf{Importance} & \textbf{Feature} & \textbf{Importance} \\
\hline
\cellcolor{blue!30}second\_relative\_integer\_time & \cellcolor{blue!30}1049 & \cellcolor{red!30}political\_government\_count & \cellcolor{red!30}96 \\
\cellcolor{blue!30}first\_relative\_integer\_time & \cellcolor{blue!30}793 & question\_count & 95 \\
avg\_word\_length & 737 & is\_same\_user & 93 \\
uppercase\_ratio & 633 & tag\_url\_count & 84 \\
char\_count & 624 & \cellcolor{red!30}trump\_specific\_total\_count & \cellcolor{red!30}84 \\
\cellcolor{blue!30}time\_diff & \cellcolor{blue!30}615 & \cellcolor{red!30}political\_candidates\_count & \cellcolor{red!30}81 \\
word\_count & 330 & \cellcolor{red!30}political\_election\_2024\_count & \cellcolor{red!30}78 \\
\cellcolor{red!30}political\_canadian\_politics\_count & \cellcolor{red!30}268 & \cellcolor{red!30}social\_citation\_patterns\_count & \cellcolor{red!30}69 \\
\cellcolor{red!30}political\_total\_count & \cellcolor{red!30}174 & \cellcolor{red!30}political\_political\_parties\_count & \cellcolor{red!30}68 \\
digit\_count & 173 & \cellcolor{red!30}social\_keywords\_total & \cellcolor{red!30}64 \\
sentence\_count & 144 & \cellcolor{red!30}trump\_specific\_trump\_names\_count & \cellcolor{red!30}59 \\
exclamation\_count & 143 & profanity\_intensity\_total\_count & 57 \\
hashtag\_count & 117 & \cellcolor{red!30}social\_engagement\_count & \cellcolor{red!30}48 \\
\hline
\end{tabular}
\caption{Top 26 feature importance scores for final LightGBM model: here we show the specific values for the Cluster 0 model, other models show similar patterns. Time features are highlighted in blue, and topic-specific keyword features are in red.}
\label{tab:featureimportancelightgbm}
\end{table*}

\subsection{Lookup Database}

We recall here that we had selected as hyperparameters the requirement that a message appear with at least 10 votes in this database, and set the winner percentage to be over 85\% (or 90\% if using the Global Fallback voting strategy).
As a result, predictions made with this route had an accuracy of 85-90\% respectively.
These highly accurate predictions were mostly for the most common actions (LIKE, FOLLOW, but also some REPLY and REPOST) due to the fact that the minimum number of 10 votes was difficult to achieve for the rare actions. This lookup route therefore contributed mainly to improving overall macro-F1 performance via the most frequently represented action categories rather the rare actions.

\subsection{LightGBM Models}
The performance for the 25 trained LightGBM models on training data is summarized in Table~\ref{tab:lightgbmmodelsf1}.

Remarkably we see that, compared to using finetuned language models, our manual text (and temporal) feature engineering approach with simpler LightGBM models outperformed by 16 macro-F1 points and 8 weighted-F1 points.
Examining the variability in class-level LightGBM performance, we see that the FOLLOW action obtained consistently good performance, the LIKE action obtained decent performance in general, and that the OTHER action was highly variable (the single model which produced an F1 score of 0.00 was the model corresponding to cluster 9 which had only 15,891 samples and very few rare actions in total).

We analyzed the feature importance of all the trained LightGBM models; we share in Table~\ref{tab:featureimportancelightgbm} the most important features for our model trained on persona cluster 0.

We noted similar feature importance trends across all 25 models: the top 10 features we always dominated by the main timestamp features, followed by several general text features such as character count or word count, and then topic-specific feature counts such as \texttt{political\_canadian\_politics\_count}.

\subsection{Rare Action Classifier}

The results for the rare action classifier are shown in Table~\ref{tab:rareactionclassifier}.

\begin{table}[h]
\centering
\begin{tabular}{lccc}
    \textbf{}  & \textbf{F1} \\
    \hline
    Majority Class Macro     & 0.07 \\
    Majority Class Weighted     & 0.34 \\
    \hline
    Text-only Model Macro     & 0.42 \\
    Text-only Model Weighted     & 0.41 \\    
    \hline
    Text + Temporal Model Macro     & \textbf{0.56} \\
    Text + Temporal Model Weighted  & \textbf{0.87} \\
\end{tabular}
\caption{Trained Rare Action Classifier performance.
 Baselines: majority class guess (UNFOLLOW, 50.4\% of 10 rare actions) and results for finetuning only the text component, i.e. ablating temporal features.
}
\label{tab:rareactionclassifier}
\end{table}

Consistent with the LightGBM feature importance results in Table~\ref{tab:featureimportancelightgbm}, we found that incorporating temporal features into the rare action classifier also significantly improved performance.

The most difficult classes were, predictably, those with very low support: UNREPOST, UNBLOCK, POST\_DELETE
--- we consistently obtained F1 scores of 0 for these three actions.

\subsection{Text Generation}

Due to the large dataset size, we randomly sampled 1,000 conversations across the 25 clusters where the user action was REPLY, and sent these messages to GPT-4.1-mini to test the ability to generate a plausible next message in the conversation. 
After our prompt engineering approaches, we achieved an average cosine similarity of 0.83 across the 1,000 conversations (min 0.70, max 0.95).

\section{Discussion}

Our results using lookup strategies alongside persona- and topic-based keywords contributed significantly to the overall classification performance of our competition submission. However, given the importance of the temporal features in improving the trained model predictions, we discuss a methodological point about the dataset and social media prediction in general.

Indeed, presumably, by far the most common reaction by most users to most social media messages is to take no action whatsoever, i.e. to just scroll past after briefly reading, but by the design of the dataset all messages were known to contain a user action. Therefore the dataset samples are a very specific subset of the entire platform and its user base---they are messages which are known to have caused at least some users to want to interact with them in some way.

In concrete terms, since the timestamp data of the messages was available and since the most common actions are LIKE and FOLLOW, it is reasonable to expect that if a user interacts with a message in under 30 seconds (for example) then their action will be a LIKE, because the very fact that they were able to respond in such a short time suggests that they are likely to be already following the message author (otherwise they would not have seen the message).

In practice, outside of the competition setting where this historical timestamp data would not be available, we believe that this notion of interaction temporal dynamics could be studied with indirect methods: for example taking into account whether or not a user already follows another, or whether they have recently refreshed their front page, could all be proxies for how quickly a user interacts with a message. We leave this as a direction for future work.

Finally, we studied user behavior in the context of Bluesky by using the persona cluster of the responding user but it would be interesting to extend the persona labeling and annotation to the user creating the original message also. This information was not available in the existing version of \textsc{BluePrint}, but it would allow further research into inter-persona interaction dynamics.

\section{Conclusion}
We presented a comprehensive hybrid approach to social media user behavior prediction that addresses the challenge of predicting both common and rare user actions on social platforms. Our methodology demonstrates that different types of user behaviors and personas require specialized modeling approaches: lookup-based prediction for historically consistent patterns (achieving 85-90\% accuracy for 22\% of samples), persona-specific tabular models for common actions (outperforming transformer models by 16 macro-F1 points), and hybrid neural architectures for rare action classification (achieving 0.56 macro-F1). The performance on extremely rare actions suggests challenges in predicting very low-frequency behaviors that may require alternative approaches such as synthetic data augmentation.

\section{Ethics Statement}
The dataset used in this paper has been made public for the SocialSim: Social-Media Based Personas shared task.
The dataset creators used publicly available Bluesky data collected in compliance with the platform’s Terms of Service \cite{bückkaeffer2025textttblueprintsocialmediauser}. To protect user privacy, data was anonymised by removing identifiable information, using relative timestamps, pseudonymizing usernames, and analyzing aggregated behaviors rather than individual actions. Users can request data removal at any time. The dataset will be shared on HuggingFace under a license restricting use to research purposes and prohibiting unethical applications. Only anonymized records are provided, supporting research on NLP techniques in social media contexts while acknowledging potential dual-use risks and emphasizing responsible AI governance.

We note that generative language models were used solely for text refinement in preparing this manuscript.

\section{Acknowledgments}
We thank the organizers of the Social Simulation with LLMs workshop at COLM 2025 for designing this very interesting challenge, and for their work in curating and publishing the Bluesky dataset.

\section{Bibliographical References}\label{sec:reference}

\bibliographystyle{lrec2026-natbib}
\bibliography{lrec2026-example}
\newpage
\appendix 
\section{Appendix}
\label{sec:appendix}
\subsection{Prompts}
\label{subsec:appendix_prompts}

\centering 
\begin{tcolorbox}[colback=white, colframe=black, title={GPT-4.1-mini system prompt}] 
You are a politically liberal human social media user. 
\end{tcolorbox} 

\centering 
\begin{tcolorbox}[colback=white, colframe=black, title={GPT-4.1-mini user prompt}] 
Below is a series of one or more messages from the social network BlueSky, which is a more liberal variant of Twitter or X.

Read the entire conversation and then generate what you think is a suitable next reply message. There is a 300 character limit, so don't write long paragraphs.

The majority of users are politically liberal, from the USA or Canada or Western Europe, and the conversation took place in 2024.

If you see any words such as @<USERNAME> it is because the data has been anonymized.

In rare cases the conversation might consist of a single user posting a multi-message thread, so if so you should try to continue in their style.

\#\# Conversation to reply to
 
\{{\texttt{conversation}}\}
\end{tcolorbox}

\end{document}